%% file: 0_main.tex
\documentclass[journal]{IEEEtran}

\usepackage{cite}
\usepackage{amsmath,amsfonts,amssymb}
\usepackage{algorithm}
\usepackage{algorithmic}
\usepackage{array}
\usepackage[caption=false,font=normalsize,labelfont=sf,textfont=sf]{subfig}
\usepackage{textcomp}
\usepackage{stfloats}
\usepackage{url}
\usepackage{verbatim}
\usepackage{graphicx}
\usepackage{extarrows}
\usepackage{makecell}
\usepackage{multicol}
\usepackage{multirow}
\usepackage{hhline}
\usepackage{bm}
\usepackage{enumitem}
\usepackage{bbding}
\usepackage{utfsym}
\usepackage{color}
\usepackage[table]{xcolor}
\usepackage{colortbl}
\usepackage[most]{tcolorbox}
\usepackage{fancyvrb}
\usepackage{booktabs}
\usepackage{balance}
\usepackage{orcidlink}
\definecolor{lightgray}{rgb}{0.88, 0.92, 0.98}
\definecolor{defblue}{rgb}{0.1843, 0.3333, 0.6}
\definecolor{defred}{rgb}{0.88, 0.2510, 0.3294}
\definecolor{lightorange}{rgb}{0.90, 0.55, 0.32}

\definecolor{bg_sect_a}{HTML}{E8F4F8}
\definecolor{bg_sect_b}{HTML}{FFF8F0}
\definecolor{bg_sect_c}{HTML}{F2F2F2}

\definecolor{exp_back}{HTML}{F4F4FB}
\definecolor{exp_sota}{HTML}{FFF3D0}
\definecolor{green1}{HTML}{EBF7F3}
\definecolor{green2}{HTML}{EFF2EB}

\newcommand{\slot}[1]{\texttt{[#1]}}

\makeatletter
\newcommand{\thickhline}{\noalign{\global\arrayrulewidth=1.5pt}\hline\noalign{\global\arrayrulewidth=0.4pt}}
\makeatother

\begin{document}

\title{FashionLens: Toward Versatile Fashion Image Retrieval via Task-Adaptive Learning}

\author{Haokun~Wen~\orcidlink{0000-0003-0633-3722},
        Xuemeng~Song~\orcidlink{0000-0002-5274-4197},~\IEEEmembership{Senior~Member,~IEEE},
        Xinghao~Xie~\orcidlink{0009-0007-1041-6420},
        Xiaolin~Chen~\orcidlink{0000-0003-4638-0603},
        Xiangyu~Zhao~\orcidlink{0000-0003-2926-4416},~\IEEEmembership{Member,~IEEE},
        and~Weili~Guan~\orcidlink{0000-0002-5658-5509},~\IEEEmembership{Member,~IEEE}
\thanks{H. Wen is with the School of Computer Science and Technology, Harbin Institute of Technology (Shenzhen), Shenzhen 518055, China, and also with the Department of Data Science, City University of Hong Kong, Hong Kong, China (e-mail: haokunwen@outlook.com).}
\thanks{X. Song is with the Department of Computer Science and Engineering, Southern University of Science and Technology, Shenzhen 518055, China, (e-mail: sxmustc@gmail.com).}
\thanks{X. Xie is with the School of Artificial Intelligence, Nanjing University, Nanjing 210023, China (e-mail: 202100161170@mail.sdu.edu.cn).}
\thanks{X. Chen is with the Institute of Data Science, National University of Singapore, Singapore (e-mail: cxlicd@gmail.com).}
\thanks{X. Zhao is with the Department of Data Science, City University of Hong Kong, Hong Kong, China (e-mail: xy.zhao@cityu.edu.hk).}
\thanks{W. Guan is with the School of Information Science and Technology, Harbin Institute of Technology (Shenzhen), Shenzhen 518055, China, and also with the Shenzhen Loop Area Institute, Shenzhen 518045, China (e-mail: honeyguan@gmail.com).}
\thanks{Corresponding authors: Xuemeng Song, Xiangyu Zhao, and Weili Guan.}
% \thanks{This work was supported by the National Natural Science
% Foundation of China under Grant 62XXXXXX.}
}

\markboth{Journal of \LaTeX\ Class Files,~Vol.~14, No.~8, August~2021}%
{Wen \MakeLowercase{\textit{et al.}}: FashionLens: Versatile Fashion Image Retrieval}

\maketitle

\begin{abstract}
Fashion image retrieval is a cornerstone of modern e-commerce systems. A unified framework that supports diverse query formats and search intentions is highly desired in practice. However, existing approaches focus on narrow retrieval tasks and do not fully capture such diversity. Therefore, in this work, we aim to develop a unified framework capable of handling diverse realistic fashion retrieval scenarios, achieving truly versatile fashion image retrieval. To establish a data foundation, we first introduce U-FIRE, a comprehensive benchmark that consolidates fragmented fashion datasets into a unified collection, supplemented by two manually curated datasets for testing generalization. Building upon this, we propose FashionLens, a unified framework based on Multimodal Large Language Models. To handle divergent matching objectives, we design a Proposal-Guided Spherical Query Calibrator that dynamically shifts query representations into task-aligned metric spaces via adaptive spherical linear interpolation. Additionally, to mitigate the optimization imbalance caused by varying task complexities and data scales, we develop a Gradient-Guided Adaptive Sampling strategy that automatically re-weights tasks based on real-time learning difficulty and the data scale prior. Experiments on U-FIRE show that FashionLens achieves state-of-the-art performance across diverse retrieval scenarios and generalizes robustly to unseen tasks. The data and code are publicly released at \url{https://github.com/haokunwen/FashionLens}.
\end{abstract}


% --- Body ---
% \IEEEpeerreviewmaketitle

\input{1_introduction.tex}

\input{2_related_work.tex}
\input{3_data.tex}

\input{4_method.tex}

\input{5_experiment.tex}
\input{6_conclusion.tex}

% \section*{Acknowledgments}
% The authors would like to thank \ldots

% --- Bibliography ---
\bibliographystyle{IEEEtran}
\bibliography{reference}

% --- Optional: author biographies (uncomment for camera-ready) ---
% \begin{IEEEbiography}[{\includegraphics[width=1in,height=1.25in,clip,keepaspectratio]{author1.jpg}}]{First A. Author}
% \end{IEEEbiography}

\end{document}

%% file: 1_introduction.tex
\section{Introduction}
\vspace{0.5em}
\begin{flushright}
    \textit{``Clothes make the man.''} \\
    \vspace{0.1em}
    --- \textup{Old Proverb}
    \vspace{-0.2em}
\end{flushright}

\IEEEPARstart{T}{he} age-old adage cited above underscores the profound significance of attire in human society, serving not merely as a functional necessity but as a pivotal medium for self-expression and identity. In the digital era, this pursuit of style has seamlessly transitioned from physical boutiques to online platforms, elevating fashion image retrieval to a cornerstone of e-commerce systems. In the fashion domain, user search demands are highly diverse, characterized by varied query formats (such as images, text, sketches, videos, or their multimodal combinations) and distinct search intentions, including finding similar, compatible, or attribute-specific items.

Early studies~\cite{SongFHYLN18, DBLP:journals/tmm/JingCGNS23, DBLP:journals/tmm/GuWSPCK19, 11488029, QMUL_Shoe_V2, Deepfashion, MovingFashion} typically developed task-specific retrieval models, leading to fragmented solutions and limited data efficiency. 
Relying on independent models for every retrieval scenario restricts the ability of a system to handle the full variety of user search demands. Consequently, there is a growing need for a unified framework that supports diverse query formats and multiple search intentions within a single architecture.
Following this direction, several existing approaches like FashionBERT~\cite{fashionbert}, Fame-ViL~\cite{famevil}, and FashionSAP~\cite{fashionsap} fine-tune vision–language pre-trained (VLP) models~\cite{clip,blip} with multiple retrieval-oriented objectives, aiming to learn more generalized representations for fashion retrieval.  Despite their effectiveness, built on VLP backbones, these methods are inherently limited in handling complex inputs, such as videos or multi-image queries. 
Consequently, they address a restricted range of tasks and fall short of the versatility required for realistic fashion retrieval scenarios.

\begin{figure}[tp]
	\includegraphics[width=\linewidth]{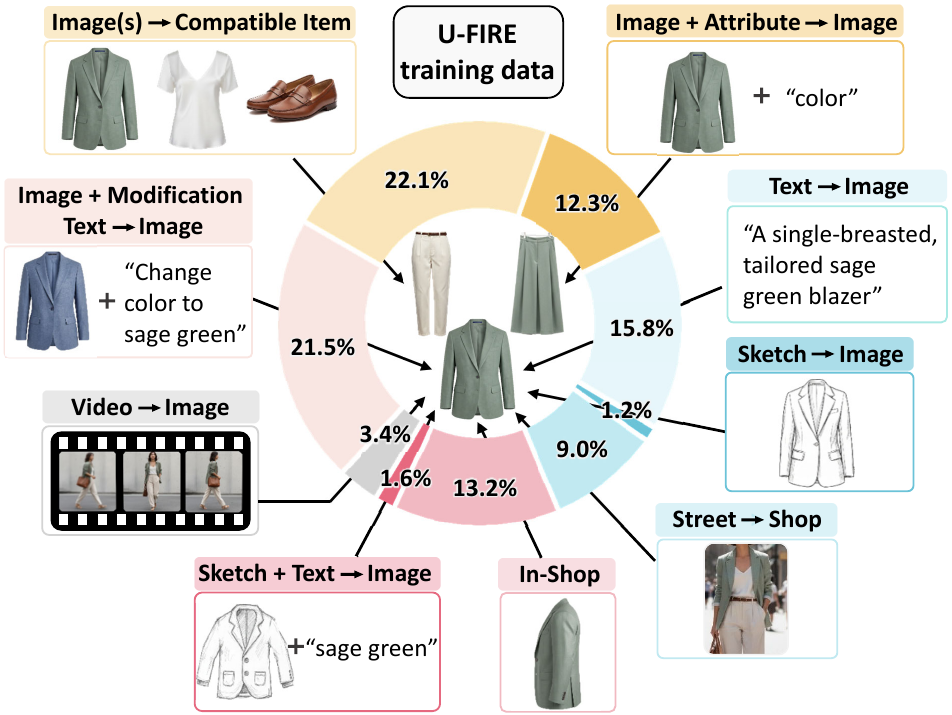}
    % \vspace{-0.5em}
	\caption{Examples of diverse fashion image retrieval tasks and the training data distribution of the constructed U-FIRE benchmark.}
	\label{fig:data}
    \vspace{-1.4em}
\end{figure}

In this paper, we aim to achieve a versatile fashion image retrieval framework that supports diverse user inputs and retrieval needs, as illustrated in Figure~\ref{fig:data}. Towards this end, inspired by recent advancements in general-domain universal multimodal retrieval~\cite{uniir, vlm2vec, VLM2VECV2}, we recognize that consolidating fragmented datasets into a unified repository is a critical first step. To address this gap in the fashion domain, we introduce
the Unified Fashion Image Retrieval \& Evaluation (U-FIRE) benchmark, which integrates $15$ fragmented fashion retrieval datasets covering $9$ existing retrieval tasks into a unified collection through instruction-following templates. The resulting benchmark comprises over $400,000$ samples, and importantly, includes two newly proposed tasks for out-of-distribution evaluation. Overall, this benchmark lays a solid foundation for advancing versatile fashion retrieval research.

Building upon this data foundation, a straightforward solution for versatile fashion image retrieval is to follow the general-domain universal multimodal retrieval paradigm: using a VLP model or Multimodal Large Language Model (MLLM) as a unified encoding backbone for both queries and targets, and fine-tuning it with contrastive learning objectives on the unified dataset. While effective in the general domain, directly transplanting it to the fashion domain remains suboptimal due to two domain-specific challenges.

\textbf{First, at the representation level, diverse retrieval intents impose divergent matching objectives, leading to significant feature interference.} 
In general-domain universal retrieval, tasks typically follow a uniform query–target matching paradigm, i.e., similarity-oriented matching. In contrast, fashion retrieval is a multi-faceted problem where the definition of  ``matching'' shifts dynamically based on user intent, ranging from capturing fine-grained visual identities in similarity matching to emphasizing abstract stylistic harmony in compatibility matching. In this context, a unified model often produces a compromised query representation that cannot perfectly satisfy each matching paradigm. This phenomenon, often characterized by negative transfer~\cite{negative1, negative2}, arises because these heterogeneous tasks place diverse semantic demands on the shared embedding space. The model is forced to settle for a semantic compromise during joint training, which ultimately undermines task-wise adaptivity. 

\textbf{Second, at the optimization level, uneven data scales and heterogeneous task complexities lead to severe training imbalance.}
Unlike general-domain data, which can often be harvested from the web, fashion data is primarily provided by online retailers, making it significantly harder to acquire. Consequently, the fashion domain exhibits more severe data imbalances than those typically encountered in general domain. In the U-FIRE benchmark, the training data distribution is highly skewed: some tasks provide ample samples, while others are sparse (Figure~\ref{fig:data}). 
Intuitively, tasks with fewer samples receive less weight under conventional scale-based sampling, which inevitably prevents them from achieving sufficient optimization. 
Moreover, retrieval tasks inherently vary in learning difficulty due to differences in query complexity and modality composition. For example, the Image+Modification Text$\to$Image task is harder to optimize than the simpler Text$\to$Image matching task. Overall, these properties prevent standard training strategies from achieving optimal performance in versatile fashion retrieval.

To address these challenges, we present FashionLens, a unified framework for versatile fashion image retrieval via task-adaptive learning. Built upon an MLLM backbone, our framework inherently supports diverse input formats including image, text, video, or multimodal combinations. 
To reconcile the divergent matching objectives within a unified model, we design a Proposal-Guided Spherical Query Calibrator (PGSQC) that operates at the representation level to resolve feature interference. By leveraging adaptive spherical interpolation, PGSQC dynamically rotates the initial query representation into a task-aligned metric space, highlighting intent-relevant features while suppressing irrelevant noise.
To mitigate optimization imbalance, we propose a Gradient-Guided Adaptive Sampling (GGAS) strategy that operates at the optimization level to mitigate training imbalance. GGAS dynamically estimates the real-time learning difficulty of each task based on gradient norms while simultaneously incorporating dataset scale as a refinement to stabilize the sampling process. This mechanism automatically re-weights tasks during training to ensure robust and balanced convergence across the entire spectrum of fashion retrieval tasks. Extensive experiments on U‑FIRE demonstrate that FashionLens achieves state‑of‑the‑art performance across diverse retrieval tasks and generalizes robustly to unseen scenarios.
Our main contributions can be summarized as follows.
\begin{itemize}[leftmargin=\parindent, labelsep=0.5em]
    \item We introduce U-FIRE, a comprehensive benchmark unifying $15$ datasets spanning $9$ existing fashion retrieval tasks, and additionally proposing $2$ new tasks for out-of-distribution (OOD) evaluation, providing a standardized testbed for versatile fashion retrieval research.
    \item We propose FashionLens, a unified MLLM-based framework for versatile fashion image retrieval, where we design a proposal-guided spherical query calibrator that dynamically modulates entangled representations into task-aligned metric spaces, to reconcile divergent matching objectives.
    \item We introduce a gradient-guided adaptive sampling strategy that leverages gradient signals to balance learning across heterogeneous tasks with uneven data scales, mitigating optimization skews for stable multi-task convergence.
    % \item We conduct extensive experiments on U‑FIRE, demonstrating that FashionLens achieves state‑of‑the‑art performance across diverse retrieval tasks and generalizes robustly to unseen scenarios. The source code and data are publicly available at \url{https://github.com/haokunwen/FashionLens}.
\end{itemize}

%% file: 2_related_work.tex
\section{Related Work}
Our work is closely in line with fashion vision-language pre-training and general-domain universal multimodal retrieval.

\subsection{Fashion Vision-Language Pre-training}
In the fashion domain, several studies~\cite{fashionbert, fashionsap, dafashion} aim to address multiple heterogeneous tasks such as retrieval, category recognition, and fashion captioning with a single model.
Representative methods like FAME-ViL~\cite{famevil} introduce three modes for different tasks and optimize the model through multi-task learning with data size proportional sampling.
FashionSAP~\cite{fashionsap} further incorporates attribute information to enhance fine-grained fashion feature representations. 
However, these models are primarily trained on image-text datasets (e.g., FashionGen~\cite{fashiongen}) or composed image retrieval datasets (e.g., FashionIQ~\cite{FashionIQ}). Such a restricted training scope fails to cover the full spectrum of fashion retrieval tasks, thereby hindering the development of versatile retrieval models in practice.
Additionally, these models mainly rely on VLP backbones~\cite{clip}, which are ill-suited for handling complex modalities such as multiple images or video clips. This limitation further constraining their applicability to broad fashion retrieval scenarios.

\subsection{General-domain Universal Multimodal Retrieval}
The field of general-domain universal multimodal retrieval (GUMR) has witnessed significant advancements in recent years. UniIR~\cite{uniir} pioneered this direction by establishing the M-BEIR benchmark, a robust data foundation that aggregates $10$ datasets across diverse domains.
While this initial work relied on VLP models such as CLIP, subsequent works like VLM2Vec~\cite{vlm2vec, VLM2VECV2}, GME~\cite{gme}, and MM-Embed~\cite{mmembed} 
have further propelled this direction by leveraging MLLMs as unified encoding backbones. Despite their strong performance in general retrieval, existing methods primarily focus on a single type of query–target matching, typically similarity-based retrieval. While sufficient for conventional tasks, this single matching paradigm is inadequate for fashion retrieval, where user intents are diverse and require divergent matching objectives, ranging from intra-category similarity retrieval~\cite{dqucir,fashiongen} to inter-category compatibility modeling~\cite{DBLP:journals/tmm/DingMWC22,DBLP:journals/tmm/LuHYCZ23}.
Additionally, existing GUMR methods typically employ a data scale–based sampling strategy. Although this approach proves effective for general-domain scenarios with relatively balanced distributions, it may fail when faced with fashion datasets that exhibit severe data imbalances and heterogeneous tasks with intrinsically different levels of training difficulty.

%% file: 3_data.tex
\begin{table*}
    \centering
\caption{Overview of the U-FIRE benchmark, which comprises $11$ tasks: nine tasks that support training, validation, and testing, and two unseen tasks reserved exclusively for evaluation.}  
\vspace{-0.5em}
\scriptsize

    \renewcommand{\arraystretch}{0.85}
    
    \resizebox{\textwidth}{!}{%
    \begin{tabular}{@{}llp{0.18\textwidth}r@{\;\;\;}r@{\;\;\;}r@{\;\;\;}r@{}}
    \toprule

        \textbf{Task}    & \textbf{Dataset}             & \textbf{Instruction} \tiny{(shown $1$ out of $4$)}    &  \textbf{Train} & \textbf{Validation} & \textbf{Test} & \textbf{Gallery}  \\
        \midrule 

         \multirow{2}{*}{1. Text$\to$Image} 

         & FashionGen~\cite{fashiongen}                 & \multirow{2}{=}[2pt]{search for images using the description text: \slot{TEXT}}              & $49,084$    &  $5,000$ & $6,854$ & $51,089$    \\
         & Shoes~\cite{Shoes} &       & $2,400$   &  $600$        & $600$ & $1,200$     \\

         \midrule
         \multirow{3}{*}{2. Sketch$\to$Image} 
         & HAIFashion~\cite{HAIFashion}                 & \multirow{3}{=}{retrieve real images that resemble the sketch}      & $1,900$    & $600$        & $600$ & $1,200$     \\
         & QMUL-Shoes-V2~\cite{QMUL_Shoe_V2}    &           & $1,600$   &  $200$        & $200$ & $400$     \\
         & ClothesV1~\cite{Clothes_V1} &  & $280$ & $100$ & $120$ & $220$ \\
         \midrule

         \multirow{1}{*}[-3pt]{3. Street$\to$Shop} & \multirow{1}{*}[-3pt]{DeepFashion2~\cite{Deepfashion2}} & {retrieve images that match the \slot{C} from the photo provided}  & \multirow{1}{*}[-3pt]{$29,187$} & \multirow{1}{*}[-3pt]{$3,497$} & \multirow{1}{*}[-3pt]{$6,037$} & \multirow{1}{*}[-3pt]{$9,369$}  \\
         \midrule 
         
         \multirow{1}{*}[-3pt]{4. In-Shop} & \multirow{1}{*}[-3pt]{DeepFashion~\cite{Deepfashion}} & {identify images that resemble the \slot{C} in the input image}  & \multirow{1}{*}[-3pt]{$43,019$} & \multirow{1}{*}[-3pt]{$5,395$} & \multirow{1}{*}[-3pt]{$5,410$} & \multirow{1}{*}[-3pt]{$10,634$}  \\
         \midrule

         \multirow{1}{*}[-3pt]{5. Video$\to$Image} & \multirow{1}{*}[-3pt]{MovingFashion~\cite{MovingFashion}} & {search for images based on the video focus on the main cloth} & \multirow{1}{*}[-3pt]{$10,909$} & \multirow{1}{*}[-3pt]{$1,294$} & \multirow{1}{*}[-3pt]{$1,290$} & \multirow{1}{*}[-3pt]{$2,608$}  \\
         \midrule 

          \multirow{1}{*}[-3pt]{6. Sketch+Text$\to$Image} & \multirow{1}{*}[-7pt]{CSTBIR~\cite{cstbir}} & {locate real images that match the sketch matching the text: \slot{TEXT}}  & \multirow{1}{*}[-7pt]{$5,158$} & \multirow{1}{*}[-7pt]{$1,500$} & \multirow{1}{*}[-7pt]{$1,500$} & \multirow{1}{*}[-7pt]{$2,632$}  \\

          \midrule 

          \multirow{5}{*}{7. Image+Modification Text$\to$Image} 
         & FashionIQ-Dress~\cite{FashionIQ}                 & \multirow{5}{=}{search for \slot{C} images matching the text modifications of the given image: \slot{TEXT}}     & $4,985$    &  $1,000 $ & $ 2,017$ & $5,650$    \\
         
         & FashionIQ-Shirt~\cite{FashionIQ} &       & $4,988 $   &  $1,000 $        & $2,038 $ & $8,215$     \\
         
    & FashionIQ-Toptee~\cite{FashionIQ} &      & $ 5,027$   &  $1,000 $        & $1,961 $ & $7,237$     \\
    
        & Shoes~\cite{Shoes} &     & $7,490$   &  $1,500$        & $1,761$ & $4,910$     \\
        
            & Fashion200K~\cite{Fashion200k} &      & $47,264$   &  $1,745$        & $2,780$ & $2,790$     \\

        \midrule

         \multirow{2}{*}{8. Image(s)$\to$Compatible Item} 
         & FashionVC~\cite{fashionvc}                 & \multirow{2}{=}[2pt]{find compatible \slot{C} go well with the given images}             & $30,718$    &  $3,858$ & $3,850$ & $7,006$    \\
         & Polyvore~\cite{polyvore} &   & $41,085$   &  $2,602$        & $4,425$ & $7,249$     \\
        \midrule 
         
         \multirow{2}{*}{9. Image+Attribute$\to$Image} 
         & FashionAI~\cite{FashionAI}                 & \multirow{2}{=}[1pt]{find images with the same \slot{ATTR} with the given image}      & $20,000$    & $3,061$        & $3,078$ & $14,400$     \\
         & DARN~\cite{DARN}    &    & $20,000$   &  $2,235$        & $2,238$ & $8,854$     \\
          \midrule
          \multicolumn{3}{c}{\textbf{Total (Tasks 1--9)}} & \textbf{325,094} & \textbf{36,187} & \textbf{46,759} & -- \\

        \midrule
        \rowcolor{exp_back}
         \multicolumn{7}{c}{\textbf{Evaluation-Only Tasks (Unseen Tasks)}} \\

         \midrule[0.2pt]

         \multirow{1}{*}[-3pt]{10. Street+Modification Text$\to$Shop} & \multirow{1}{*}[-3pt]{DeepFashion2~\cite{Deepfashion2}} & {locate \slot{C} images per the text changes to the given image}  & \multirow{1}{*}[-3pt]{$-$} & \multirow{1}{*}[-3pt]{$-$} & \multirow{1}{*}[-3pt]{$1,000$} & \multirow{1}{*}[-3pt]{$1,765$}  \\
         \midrule 

         \multirow{1}{*}[-3pt]{11. Image(s)+Text$\to$Compatible Item} & \multirow{1}{*}[-3pt]{Polyvore~\cite{polyvore}} & {find compatible \slot{TEXT} \slot{C} go well with the given images}  & \multirow{1}{*}[-3pt]{$-$} & \multirow{1}{*}[-3pt]{$-$} & \multirow{1}{*}[-3pt]{$1,000$} & \multirow{1}{*}[-3pt]{$2,302$}  \\
         
         \bottomrule
    \end{tabular}
    }
    \vspace{0.5pt}
    \flushleft{\footnotesize \textit{Note:} \slot{C}, \slot{ATTR}, and \slot{TEXT} represent category, attribute, and text description placeholders in the instruction templates, respectively.}
    \label{tab:dataset}
    \vspace{-1.2em}
\end{table*}

\section{U-FIRE Benchmark}
In this section, we detail our constructed U-FIRE benchmark, which not only consolidates $15$ publicly available fashion-domain datasets covering $9$ core fashion image retrieval tasks into a unified instruction-augmented format, but also introduces $2$ new tasks with manually curated datasets to support the out-of-distribution (OOD) generalization.

\subsection{Instruction Annotation} 
In conventional single-task retrieval, the retrieval objective has been fixed and implicitly embedded within the system, e.g., similarity matching or compatibility matching. Accordingly, users typically provide content-based queries, such as pure text descriptions or images, for target retrieval. However, versatile fashion image retrieval aims to handle a diverse spectrum of retrieval objectives within a single unified framework. Under this setting, raw content alone is often insufficient because the same content can correspond to different retrieval goals, e.g., similar item retrieval or compatible item retrieval.
Therefore, following~\cite{uniir}, we employ natural language instructions to explicitly specify the search intent. For each task, we manually curate a set of four instruction templates, ensuring they are syntactically distinct yet semantically equivalent. We then randomly assign one of these templates to each sample. This strategy prevents the model from overfitting to fixed sentence patterns, thereby fostering linguistic robustness.
 
By consolidating $15$ datasets, we finally obtain over $400$k samples, including $325$k for training, $36$k for validation, and $46$k for testing. Each sample in our benchmark is a triplet in the form of $\langle$raw query content, search instruction, target image$\rangle$. In addition, for each dataset, we construct a dedicated gallery image set for evaluation. Notably, to ensure representative coverage of item retrieval and benchmark quality, before instruction annotation, we first perform data filtering, where we only retain images from primary apparel categories, including \textit{Tops}, \textit{Bottoms}, \textit{Full-Body Items}, and essential \textit{Accessories} (e.g., hats, scarves, socks), while excluding auxiliary categories such as jewelry and bags, which are often small in visual scale or sparsely distributed across the datasets. 
\begin{figure}[tp]
	\includegraphics[width=\linewidth]{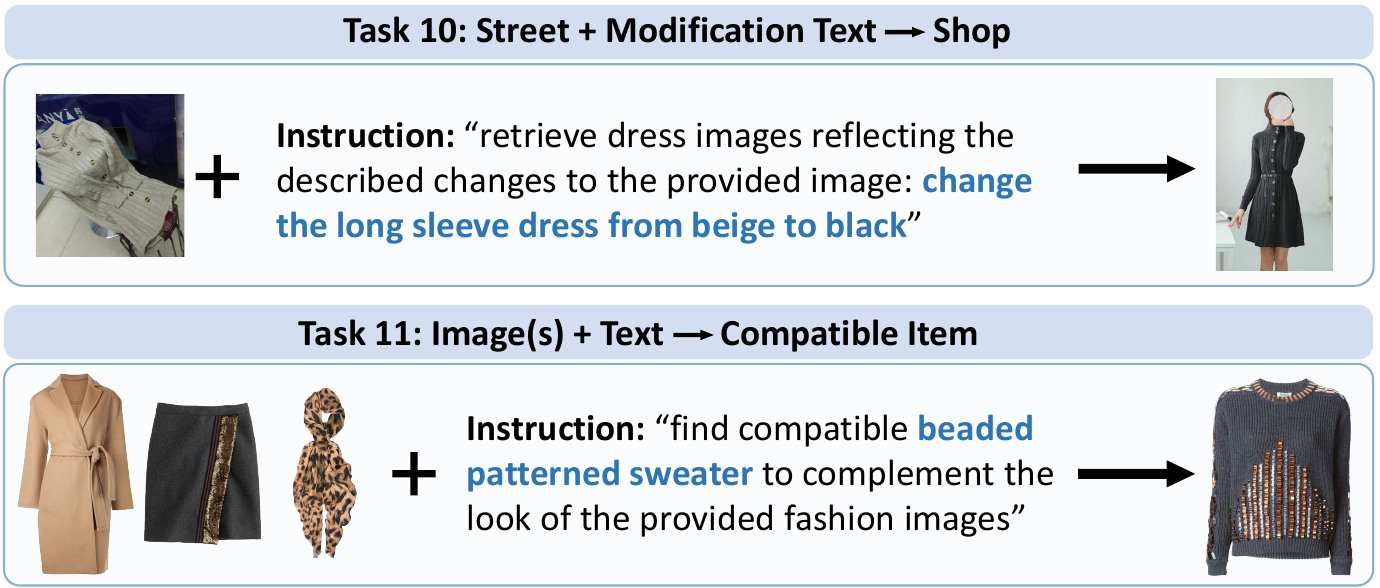}
    \vspace{-1em}
	\caption{Examples of the two unseen evaluation tasks. Blue text denotes the modification text for Task 10 and concise attribute descriptions for Task 11.}
	\label{fig:evaluation_task}
    \vspace{-1em}
\end{figure}

\subsection{Unseen Tasks for OOD Evaluation}
Although the unified dataset provides test sets for diverse tasks, their corresponding training data are used during model optimization. As a result, evaluation on these test sets primarily reflects the model's multi-task learning capability rather than its ability to generalize to unseen tasks. To assess such generalization, U-FIRE introduces two unseen tasks for out-of-distribution evaluation, as illustrated in Figure~\ref{fig:evaluation_task}. Both tasks involve multimodal queries and require complex reasoning, thereby also evaluating the model's ability to handle challenging real-world scenarios. Each unseen task is accompanied by a manually curated dataset. 

\begin{itemize}
    \item \textbf{Street+Modification Text$\to$Shop (Task 10):} 
    This task requires retrieving specific shop images based on street photos plus natural language modifications. It essentially combines the objectives of Task 3 (Street$\to$Shop) and  Task 7 (Image+Modification Text$\to$Image), reflecting a high-value real-world scenario in which users search for online products using a street-captured photo together with a textual modification description. Notably, this task differs from Image+Modification Text$\to$Image (Task 7), where both the query and target images originate from the same domain (i.e., shop images). Specifically, we curate image pairs from the DeepFashion2~\cite{Deepfashion2} test set in which street and shop images share the same item identity but differ in certain attributes (e.g., color, material, or pattern). We then manually annotate corresponding modification texts to describe these discrepancies.
    
    \item \textbf{Image(s)+Text$\to$Compatible Item (Task 11):}  
    This task aims to retrieve items that are not only stylistically compatible with a given visual context but also satisfy explicit textual constraints provided by the user. Compared with conventional compatibility matching, this setting more closely reflects realistic scenarios, as multiple compatible choices may exist and users often impose additional attribute-level requirements (e.g., color, style, or functionality) to express their personalized preferences. Essentially, this task integrates the objectives of Task 8 (Image(s)$\to$Compatible Item) and Task 1 (Text$\to$Image).  The data is derived from the Polyvore~\cite{polyvore} test split, where we retain the original visual query contexts and generate concise attribute descriptions for target items using Qwen3-VL-8B~\cite{qwen3vl}, followed by careful human verification to ensure accuracy and consistency.
\end{itemize}

%% file: 4_method.tex
\section{FashionLens}
In this section, we first present the problem formulation and introduce the standard MLLM-based query and target encoding paradigm. We then detail the two key components of FashionLens: the Proposal-Guided Spherical Query Calibrator (PGSQC), designed to mitigate feature interference, and the Gradient-Guided Adaptive Sampling (GGAS), which alleviates optimization imbalance across tasks. Finally, we describe the overall training objectives.

\begin{figure*}[htp]
\centering
	\includegraphics[width=0.82\linewidth]{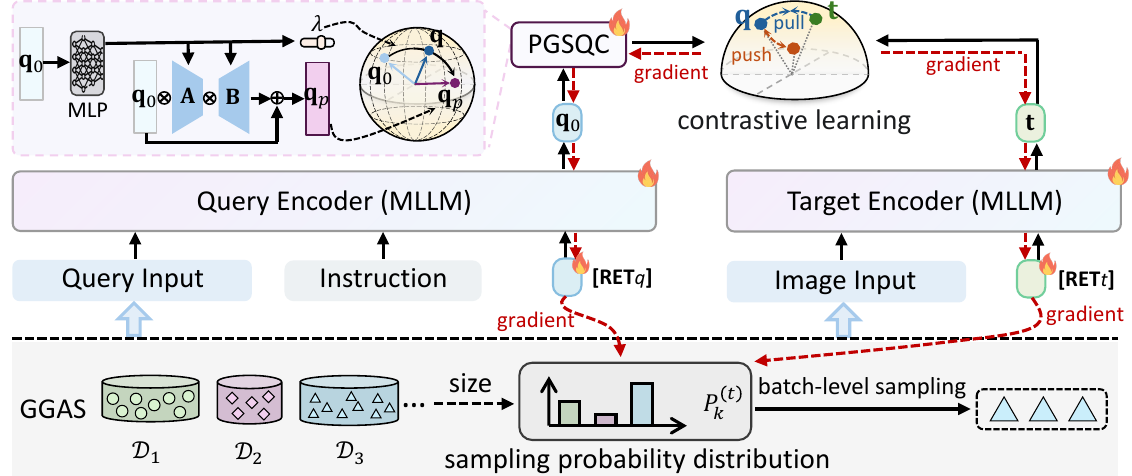}
    \vspace{-0.2em}
	\caption{Illustration of FashionLens, which leverages MLLMs as the backbone. It includes two key novel components: 1) Proposal-Guided Spherical Query Calibrator (PGSQC) and 2) Gradient-Guided Adaptive Sampling (GGAS).}
    \vspace{-0.5em}
	\label{fig:framework}
\end{figure*}

\subsection{Problem Formulation}
Formally, we define versatile fashion retrieval as a ranking problem based on intention-aware queries. Let $\mathcal{D} = \{\mathcal{D}_k\}_{k=1}^K$ be a collection of training datasets covering $O$ heterogeneous retrieval tasks, where $K \ge O $ since a single task may encompass multiple datasets. Each sample in $\mathcal{D}$ follows a unified format $(Q, T)$, where $Q = \langle R_q, I_q \rangle$ represents an intention-aware query. Here, $R_q$ represents the \textbf{raw query content}, i.e., the pure reference content without any explicit specification of the retrieval intent, which may be in a single modality (e.g., an image or text) or a combination of multiple modalities. The instruction $I_q$ denotes the \textbf{search instruction}, which is a natural language description that explicitly specifies the retrieval intent. $T$ denotes the ground-truth \textbf{target image} corresponding to the intention-aware query.

Based on the unified dataset $\mathcal{D}$, the goal is to learn a unified scoring function
$S: \mathcal{Q} \times \mathcal{G} \rightarrow \mathbb{R}$,
which effectively measures the relevance between an arbitrary intention-aware query $Q\in \mathcal{Q}$ and a candidate image $X \in \mathcal{G}$, assigning higher scores to more relevant images. $\mathcal{G}$ denotes the gallery image set. 

\subsection{MLLM-based Query/Target Encoding}
Inspired by the success of general-domain universal multimodal retrieval methods~\cite{gme,vlm2vec,VLM2VECV2,mmembed}, we adopt an MLLM as the architectural backbone. Owing to its strong ability to process heterogeneous modalities and perform semantic reasoning, this backbone enables FashionLens to support diverse query formats within a unified framework. Specifically, we append two learnable special tokens, denoted as $[\text{RET}_q]$ and $[\text{RET}_t]$, to the intention-aware query token sequence (including the raw query content and search instruction) and its corresponding target image token sequence, for aggregating their semantic information, respectively. Formally, we regard the final-layer hidden states of the $[\text{RET}_q]$ and $[\text{RET}_t]$ tokens as the query and target representations:
\begin{equation}
\left\{
\begin{aligned}
    \mathbf{q}_0 &= \text{MLLM}(R_q, I_q, [\text{RET}_q]), \\
    \mathbf{t} &= \text{MLLM}(T, [\text{RET}_t]),
\end{aligned}
\right.
\end{equation}
where $\mathbf{q}_0\in \mathbb{R}^{D}$ and $\mathbf{t}\in \mathbb{R}^{D}$ denote the initial query and target representations, respectively. Both $\mathbf{q}_0$ and $\mathbf{t}$ are $L_2$ normalized to reside on a unit hypersphere.

\subsection{Proposal-Guided Spherical Query Calibrator}

Fashion retrieval involves heterogeneous tasks with divergent matching objectives. Consequently, the jointly optimized model tends to learn a generic representation $\mathbf{q}_0$ that compromises across tasks, causing feature interference where dominant features of one task may overshadow the critical signals of another. Furthermore, natural language-based search instructions provide relatively weak conditioning compared to high-dimensional visual features, making them insufficient to independently reorient the representation toward task-optimal directions.

Given the lack of explicit supervision indicating what features are relevant to the specific intention, completing the calibration of the query representation in a single step is challenging and not robust. Therefore, we propose a Proposal-Guided Spherical Query Calibrator (PGSQC) for adapting the initial query representation to the specific search intention in a cautious and conservative manner. The core idea is to first generate an intention-oriented adaptation proposal, serving as a directional probe that highlights potential intention-relevant features while suppressing redundant signals. This probe provides a candidate direction for adaptation, but is not directly used as the final representation. Instead, we perform adaptive spherical linear interpolation (Slerp) between the original representation $\mathbf{q}_0$ and this probe, producing a robust intention-aware query that balances the initial MLLM-generated representation with the proposed adaptation.

\subsubsection{Intention-Oriented Adaptation Proposal}
Regarding the adaptation proposal generation, we introduce a pair of learnable low-rank matrices $\mathbf{A} \in \mathbb{R}^{D \times d}$ and $\mathbf{B}\in \mathbb{R}^{d \times D}$ according to the information bottleneck principle, where the down-projection $\mathbf{A}$ maps $\mathbf{q}_0$ into a compact latent space, emphasizing intention-relevant components, and the up-projection $\mathbf{B}$ reconstructs the latent representation back to the original space. The intention-oriented adaptation proposal  $\mathbf{q}_{p}\in \mathbb{R}^{D} $ is then computed as follows:
\begin{equation}
\mathbf{q}_{p} = \text{Norm}\left(\mathbf{q}_0 + \mathbf{q}_0 \mathbf{A} \mathbf{B}\right),
\end{equation} 
where $\text{Norm}(\cdot)$ denotes $\ell_2$ normalization, ensuring that $\mathbf{q}_p$ lies on the same unit hypersphere as $\mathbf{q}_0$. The residual connection in $\mathbf{q}_0 + \mathbf{q}_0 \mathbf{A}\mathbf{B}$ preserves the core semantics of the original MLLM representation while allowing the low-rank adapter to focus on learning ``what to change'', avoiding uncontrolled deviations that could distort identity-relevant features. 

Considering that different queries, even within the same retrieval task, may emphasize distinct visual attributes or stylistic cues, we dynamically generate the low-rank projection matrices $\mathbf{A}$ and $\mathbf{B}$ conditioned on the initial query representation $\mathbf{q}_0$ using a Multi-Layer Perceptron (MLP). Details of this query-adaptive parameterization are provided at the end of this subsection, since these parameters are jointly learned with the interpolation coefficient that controls the query adaptation strength in the following module.

\subsubsection{Adaptive Spherical Linear Interpolation} 
As mentioned previously, due to the lack of explicit supervision, the proposal derived from the low-rank adapter may be unreliable. Therefore, we further combine the original query $\mathbf{q}_0$ and the adaptation proposal to obtain the final intention-aware query representation. 
Fashion retrieval typically computes the dot product of normalized query and target embeddings. Since all embeddings are unit vectors, this is equivalent to cosine similarity, which reflects the angular proximity between the query and target representations. Accordingly, adapting a query representation can be interpreted as rotating it along the hypersphere toward a target representation. Therefore, we perform spherical linear interpolation (Slerp) between $\mathbf{q}_0$ and $\mathbf{q}_p$ to obtain the final representation, rather than using traditional linear interpolation in Euclidean space.

Notably, standard Slerp employs a fixed interpolation coefficient $\lambda$, which overlooks the varying degrees of semantic deviation inherent in the compromised representation $\mathbf{q}_0$ and risks either under-correcting misaligned queries or over-distorting well-aligned ones. In practice, for queries where $\mathbf{q}_0$ already captures dominant intent-relevant features, a small $\lambda$ is sufficient. However, when the representation suffers from severe feature interference, a larger $\lambda$ becomes essential to rigorously suppress noise and reorient the embedding into the correct task-aligned metric space.

Toward this end, we propose adaptive Slerp, where the interpolation coefficient $\lambda(\mathbf{q}_0)$ is query-dependent and controls the degree to which each query follows the intention-oriented proposal. Formally, the adaptive Slerp is defined as:
\begin{equation}
\begin{split}
\mathbf{q} &= \mathrm{Slerp}(\mathbf{q}_0, \mathbf{q}_p, \lambda(\mathbf{q}_0)) \\
&= \frac{\sin\!\big((1-\lambda(\mathbf{q}_0))\Omega\big)}{\sin\Omega}\mathbf{q}_0
+ \frac{\sin\!\big(\lambda(\mathbf{q}_0)\Omega\big)}{\sin\Omega}\mathbf{q}_p,
\end{split}
\label{slerp}
\end{equation}
where $\Omega = \operatorname{arccos} \left( \mathbf{q}_0^\top \mathbf{q}_{p} \right)$ is the angle between $\mathbf{q}_0$ and $\mathbf{q}_p$ on the hypersphere. Since both representations are normalized, their dot product equals the cosine of the angle between them. The interpolation coefficient $\lambda(\mathbf{q}_0) \in (0,1)$ determines the degree to which the adapted query aligns with the intention-oriented proposal.  It is learnable based on the initial query representation $\mathbf{q}_0$, analogous to the way the low-rank adapter parameters $\mathbf{A}$ and $\mathbf{B}$ are learned.

\subsubsection{Adaptive Joint Parameterization} 
Rather than learning the low-rank projection matrices $\mathbf{A}$, $\mathbf{B}$, and the interpolation coefficient $\lambda$ independently, we jointly predict them conditioned on the initial query representation $\mathbf{q}_0$ using an MLP:
\begin{equation}
(\mathbf{A}, \mathbf{B}, \lambda) = \mathrm{MLP}(\mathbf{q}_0).
\label{mlp}
\end{equation}
Intuitively, the low-rank projections $\mathbf{A}$ and $\mathbf{B}$ define a candidate direction toward an intention-oriented proposal representation, while the interpolation coefficient $\lambda$ controls how strongly the model follows this direction. By jointly learning $(\mathbf{A}, \mathbf{B}, \lambda)$, the model can effectively share information across these parameters, allowing a coherent coordination of both the adaptation direction and the degree of query refinement.

\textit{Regularization.} To ensure the expressiveness and stability of the intention-oriented proposal generation, we impose two structural constraints on the low-rank bottleneck matrices. First, since the down-projection matrix $\mathbf{A}$ defines a compact latent subspace for extracting intention-relevant information from the original query representation, we impose an orthogonality constraint on $\mathbf{A}$ to encourage the $R$ latent dimensions to capture complementary and non-redundant semantic factors, as follows:
\begin{equation}
    \mathcal{L}_{ortho} = \left\| \mathbf{A}^\top \mathbf{A} - \mathbf{I} \right\|_{F}^{2}.
\end{equation}
We do not apply this constraint to the up-projection matrix $\mathbf{B}$, as it is only responsible for reconstructing the representation back to the original space rather than defining the latent intention subspace.

Meanwhile, to regularize the overall magnitude of the query adaptation and stabilizes training, we apply a Frobenius norm penalty to both projection matrices:
\begin{equation}
    \mathcal{L}_{reg} = \| \mathbf{A} \|_F^2 + \| \mathbf{B} \|_{F}^{2}.
\end{equation}

\subsection{Gradient-Guided Adaptive Sampling}
Existing multi-task retrieval frameworks typically rely on static sampling, such as heuristics based on dataset size~\cite{famevil,gme,mmembed}. This reliance on dataset scale can lead to under-optimization for low-resource tasks, as gradients are often dominated by large-scale datasets. More critically, it overlooks the intrinsic differences in learning difficulty across tasks. In fashion retrieval, query complexity and modality composition vary drastically across different tasks. For example, Image+Modification Text$\to$Image requires sophisticated understanding of fine-grained visual changes, making it inherently harder to optimize than straightforward Text$ \to$Image retrieval.
Therefore, we propose Gradient-Guided Adaptive Sampling (GGAS), which leverages gradient-based difficulty signals to guide task prioritization. 
Notably, to prevent the model from over-emphasizing tasks with limited training data, where gradient estimates tend to exhibit higher variance, we introduce the dataset scale as a calibration factor to stabilize the sampling process.

\subsubsection{Difficulty Quantification via Retrieval Token Gradients}
Directly quantifying task-level learning difficulty via full-parameter gradients is computationally prohibitive, especially for large-scale MLLMs. Instead, we leverage the learnable retrieval tokens ($[\text{RET}_q]$, $[\text{RET}_t]$) as a compact and task-sensitive proxy for optimization demand. These tokens form a structural bottleneck for multimodal aggregation and query–target alignment, making their gradient norms a reliable indicator of real-time learning difficulty.

Formally, let $\mathcal{L}_{k}^{(t)}$ denote the loss for task $k$ at training step $t$. We quantify the {instantaneous learning difficulty} of task $k$ with the summation of the gradient magnitude of the two learnable tokens as follows:
\begin{equation}
d_k^{(t)} =
\left\|\nabla_{[\mathrm{RET}_q]} \mathcal{L}_k^{(t)}\right\|_2
+
\left\|\nabla_{[\mathrm{RET}_t]} \mathcal{L}_k^{(t)}\right\|_2 .
\end{equation}

Since this instantaneous signal can be noisy due to mini-batch stochasticity, we maintain a smoothed difficulty estimate $G_k^{(t)}$ using an Exponential Moving Average (EMA) as follows:
\begin{equation}\label{ema}
G_k^{(t)} = \alpha G_k^{(t-1)} + (1 - \alpha)\, d_k^{(t)},
\end{equation}
where $\alpha \in [0,1)$ is a smoothing coefficient.
A larger $G_k^{(t)}$ indicates greater difficulty in aligning query and target representations for task $k$, and warrants increased sampling during training.

\subsubsection{Size-Aware Probability Refinement}
Relying solely on gradient-based task difficulty may bias the selection towards low-resource tasks. This is because tasks with limited training data often yield high-variance gradient estimates due to insufficient data diversity. Consequently, they may exhibit inflated gradient norms not because they are intrinsically more complex, but because of statistical instability.
To mitigate this issue, we incorporate the dataset size into the task sampling score, ensuring that tasks with more data are adequately represented while still prioritizing tasks that are currently difficult to optimize. 
Formally, the sampling score $S_k^{(t)}$ for task $k$ at training step $t$ is defined as:
\begin{equation}\label{gamma}
S_k^{(t)} = \exp\!\left(
\frac{G_k^{(t)}}{\eta}
+ \gamma \log N_k
\right),
\end{equation}
where $N_k$ denotes the size of the $k$-th dataset.
The first term reflects the gradient-based learning difficulty, while the second term introduces a sublinear dataset-scale prior with $\gamma \in (0,1)$.
In practice, we set $\gamma=0.5$ to achieve square-root scaling. % without allowing large datasets to dominate training.
The temperature parameter $\eta$ controls the sensitivity of the sampling distribution to difficulty variations.

To ensure that every task receives at least a minimal chance of being sampled, we add a small constant to the raw sampling scores before normalizing them into a probability distribution as follows,
$\mathcal{P}^{(t)} = \{P_1^{(t)}, \ldots, P_K^{(t)}\}$ using a minimum guarantee mechanism:
\begin{equation} \label{eps}
P_k^{(t)} = \frac{\tilde{P}_k^{(t)}}{\sum_j \tilde{P}_j^{(t)}}, \quad \text{where} \quad \tilde{P}_k^{(t)} = \max \left(\frac{S_k^{(t)}}{\sum_j S_j^{(t)}}, \epsilon\right). 
\end{equation}
Here, $\epsilon$ defines a minimum sampling probability to ensure that no task is completely neglected. 
Consequently, even tasks with low difficulty or limited data remain actively represented during training, sustaining informative gradient signals and preventing under-training of task-specific representations. 

To implement task-level sampling efficiently, we adopt batch-level sampling rather than instance-level sampling. Specifically, at each training step $t$, we draw a mini-batch of $N$ samples exclusively from the task $k^{(t)}_*$ with the highest sampling probability, i.e., $k^{(t)}_* = \arg\max_k P_k^{(t)}$.
This design not only simplifies the sampling procedure but also avoids trivial cross-task negatives, ensuring that the contrastive signals in the denominator of the subsequent InfoNCE-based optimization objective remain informative within each mini-batch.

\subsection{Training Objectives}

For optimization, we adopt the widely used InfoNCE loss. For a mini batch of size $N$, the retrieval objective is formulated as follows:
\begin{equation}
\mathcal{L}_{ret} = -\frac{1}{N}\sum_{i=1}^{N} \log \frac{\exp(\mathbf{q}_{i}^\top  \mathbf{t}_{i} / \tau)}{\sum_{j=1}^{N} \exp(\mathbf{q}_{i}^{\top} \mathbf{t}_j / \tau)},
\label{infonce}
\end{equation}
where $\tau$ is the temperature parameter. 

Combining with the regularization losses, we have the final loss function as follows:
\begin{equation}
\mathcal{L}
= \mathcal{L}_{ret}
+ \beta_1 \mathcal{L}_{ortho}
+ \beta_2 \mathcal{L}_{reg}, 
\label{finalloss}
\end{equation}
where $\beta_1$ and $\beta_2$ are balancing coefficients.

%% file: 5_experiment.tex
\begin{table*}[htbp]
\centering
\setlength{\tabcolsep}{2pt}

\setlength{\extrarowheight}{3pt} 
\caption{Performance Comparison. The metric used is $\mathbf{mR = (R@1 + R@5 + R@10)/3}$ for all tasks. Methods are classified into General-domain Universal Multimodal Retrievers (GUMR), Fashion-Domain VLP Models (Fashion-VLP), and Fine-tuned Models. The best results are highlighted with\colorbox{bg_sect_b}{\phantom{----}}. The symbol ``-'' denotes the input modality (multi-image or video) is not supported. }
% \vspace{-0.5em}
\label{tab:main_results}

\resizebox{\textwidth}{!}{%
\begin{tabular}{ll|cccc|ccc|ccc}
\thickhline

\multirow{2}{*}{\textbf{Task}} & \multirow{2}{*}{\textbf{Dataset}} & \multicolumn{4}{c|}{\textbf{GUMR (general-domain)}} & \multicolumn{3}{c|}{\textbf{Fashion-VLP (fashion-domain)}} & \multicolumn{3}{c}{\textbf{Fine-tuned Models}} \\

\hhline{~~|----|---|---|}

 & & UniIR & VLM2Vec-V2 & MM-Embed & GME & Fame-ViL & FashionSAP & DA-Fashion & UniIR$^*$ & Qwen3-VL & FashionLens \\
\hline
\hline

% Task 1
\multirow{2}{*}{1. Text$\to$Image} & FashionGen & $17.39$ & $33.93$ & $25.86$ & $37.80$ & $13.38$ & $48.06$ & \cellcolor{bg_sect_b} $\textbf{60.34}$ & $54.90$  & $46.41$ & $59.45$ \\
 & Shoes & $33.84$ & $39.72$  & $37.56$ & $45.61$ & $29.33$ & $14.72$ & $14.00$ &  $39.00$ & $44.17$ &  \cellcolor{bg_sect_b} $\textbf{52.06}$ \\
\hline 

% Task 2
\multirow{3}{*}{2. Sketch$\to$Image} & HAIFashion & $36.11$ & $47.44$ & $56.05$ & $53.33$ & $20.56$ & $10.28$ & $11.17$ & $83.67$  & $54.84$ & \cellcolor{bg_sect_b} $\textbf{92.17}$ \\
 & QMUL-Shoes-V2 & $8.17$ & $1.17$ & $1.17$ & $1.67$ & $1.33$ & $1.33$ & $1.00$ &  $50.67$ & $16.33$ & \cellcolor{bg_sect_b} $ \textbf{54.50}$ \\
 & ClothesV1 & $61.67$ & $66.39$ & $67.50$ & $73.06$ & $37.22$ & $13.89$ & $20.00$ & $77.50$  & $76.39$ & \cellcolor{bg_sect_b} $\textbf{93.61}$  \\
\hline

% Task 3
3. Street$\to$Shop & DeepFashion2 & $19.64$ & $21.54$ & $32.14$ & $32.19$ & $15.86$ & $8.39$ & $4.52$ &  $76.01$ & $72.65$ & \cellcolor{bg_sect_b} $\textbf{79.26}$  \\
\hline

% Task 4
4. In-Shop & DeepFashion & $52.95$ & $58.37$ & $59.34$ & $63.68$ & $41.31$ & $43.45$ & $52.26$ & $66.13$ & $65.85$ & \cellcolor{bg_sect_b} $\textbf{66.94}$ \\
\hline

5. Video$\to$Image & MovingFashion & - & $36.90$ & - & $44.51$ & - & - & - & - & $78.44$ & \cellcolor{bg_sect_b} $\textbf{84.16}$ \\
\hline

6. Sketch+Text$\to$Image & CSTBIR & $2.85$ & $4.69$ & $5.18$ & $5.11$ & $3.93$ & $0.64$ & $0.36$ & $8.53$ & $7.84$ & \cellcolor{bg_sect_b} $\textbf{11.40}$ \\
\hline

\multirow{5}{*}{7. Image+Modification Text$\to$Image} 
 & FashionIQ-Dress & $21.78$ & $7.73$ & $22.61$ & $27.22$ & $21.04$ & $2.28$ & $0.96$ & $16.97$  & $20.97$ &  \cellcolor{bg_sect_b} $\textbf{27.37}$ \\
 & FashionIQ-Shirt & $28.18$ & $16.78$ & $28.49$ & \cellcolor{bg_sect_b} $\textbf{36.88}$ & $26.68$ & $3.29$ & $1.47$ & $19.15$  & $28.90$ &  $36.10$ \\
 & FashionIQ-Toptee & $29.34$ & $13.82$ & $29.36$ & $34.68$ & $28.30$ & $3.31$ & $1.41$ &  $21.96$ & $29.73$ & \cellcolor{bg_sect_b} $\textbf{37.77}$ \\
 & Shoes & $22.56$ & $18.27$ & $27.52$ & $33.77$ & $15.60$ & $7.36$ & $5.53$ & $33.33$  & $33.35$ & \cellcolor{bg_sect_b} $\textbf{43.10}$  \\
 & Fashion200K & $29.77$ & $30.78$ & $34.18$ & $44.24$ & $13.61$ & $11.06$ & $8.49$ &  $37.95$ & $43.86$ & \cellcolor{bg_sect_b} $\textbf{48.01}$ \\
\hline

% Task 5
\multirow{2}{*}{8. Image(s)$\to$Compatible Item} & FashionVC & $0.76$ & $0.62$ & $0.55$ & $2.21$ & $0.62$ & $0.37$ & $0.27$ & $3.79$ & $5.75$ & \cellcolor{bg_sect_b} $\textbf{5.90}$ \\
 & Polyvore & - & $1.12$ & $1.81$ & $3.37$ & - & - & - & - & $9.90$ &  \cellcolor{bg_sect_b} $\textbf{11.12}$ \\
\hline

\multirow{2}{*}{\makecell[l]{9. Image+Attribute$\to$Image}} 
 & FashionAI & $45.79$ & $42.86$ & $48.21$ & $53.87$ & $43.40$ & $30.91$ & $29.61$ &  $71.15$ & $70.58$ &  \cellcolor{bg_sect_b} $\textbf{76.31}$\\
 & DARN & $22.49$ & $23.93$ & $24.38$ & \cellcolor{bg_sect_b} $\textbf{25.54}$ & $21.39$ & $18.98$ & $17.22$ & $19.36$  & $21.46$ &  $20.94$ \\
\hline 
        \rowcolor{exp_back}
         \multicolumn{12}{c}{\textbf{Out-of-Distribution Performance}} \\
         \hline
10. Street+Modification Text$\to$Shop & DeepFashion2 & $22.77$ & $23.20$ & $28.57$ & $26.03$ & $10.97$ & $4.60$ & $4.52$ & $23.37$ & $41.80$ & \cellcolor{bg_sect_b}$\textbf{66.27}$ \\
\hline

\makecell[l]{11. Image(s)+Text$\to$Compatible Item} & Polyvore & -& $28.30$ & $45.83$ & $63.13$ & - & - & - & - & $57.83$ & \cellcolor{bg_sect_b}$\textbf{77.70}$ \\
\hline
\hline

\multicolumn{2}{c|}{\textbf{Average (Commonly Supported)}} & $26.83$ & $26.54$ & $31.10$ & $35.11$ & $20.27$ & $13.11$ & $13.71$ & $41.38$ & $40.05$ & \cellcolor{bg_sect_b} $\textbf{51.24}$ \\
\multicolumn{2}{c|}{\textbf{Average (All)}} & - & $25.88$ & - & $35.40$ & - & - & - & - & $41.35$ & \cellcolor{bg_sect_b}$\textbf{52.21}$ \\
\thickhline
\end{tabular}%
}

\vspace{6pt} 

\parbox{\textwidth}{
    \footnotesize 
    \textbf{Average (Commonly Supported)}: The
    average performance over commonly supported datasets (i.e, excluding MovingFashion in Task 5 and Polyvore in Tasks 8 \& 11).  \quad \\
    \textbf{Average (All)}: The
    average performance over all datasets covering both training and evaluation tasks.
}
\vspace{-0.2em}
\end{table*}

\section{Experiment}
In this section, we first detail the experimental settings and then present the experimental results along with corresponding analyses.
\subsection{Experimental Settings}
We adopt Qwen3-VL-4B~\cite{qwen3vl} as the MLLM backbone due to its optimal balance between efficiency and effectiveness. The dimensions for the low-rank matrices $\mathbf{A}$ and $\mathbf{B}$ are set to $D=2560$ and $d=32$. The EMA smoothing coefficient $\alpha$ in Eqn. (\ref{ema}) is set to $0.9$, while the scaling factors $\gamma$ and $\eta$ in Eqn. (\ref{gamma}) are set to $0.5$ and $1.0$, respectively. The minimum sampling probability $\epsilon$ in Eqn. (\ref{eps}) is fixed at $0.02$. For the objective function in Eqn. (\ref{finalloss}), the balancing coefficients $\beta_1$ and $\beta_2$ are set to $10^{-2}$ and $10^{-4}$, respectively. The model is fine-tuned via LoRA~\cite{lora} using the AdamW optimizer~\cite{adamw} with a learning rate of $10^{-4}$ for a total of $5$ epochs, and the LoRA rank is set to $8$. We implement a warm-up strategy for the sampling mechanism. Specifically, we use the uniform sampling over different datasets during the first epoch to exclusively track and update gradient EMA statistics, whereas the difficulty-aware sampling is activated from the second epoch onwards. Additionally, we utilize the GradCache~\cite{gradcache} strategy to expand the batch size to $64$. All experiments are conducted on two NVIDIA H20 GPUs.

\subsection{Performance Comparison}
For performance comparison, we adopt three categories of state-of-the-art models as baselines:
\begin{itemize}[leftmargin=\parindent, labelsep=0.5em]
    \item \textbf{General-domain Universal Multimodal Retrievers (GUMR):} 
    This group of models are trained on large-scale general-domain retrieval datasets, including UniIR~\cite{uniir} (CLIP-L/14 backbone with score-level fusion), VLM2Vec-V2~\cite{VLM2VECV2} (Qwen2-VL-7B backbone), MM-Embed~\cite{mmembed} (LLaVA-NeXT backbone), and GME~\cite{gme} (Qwen2-VL-7B backbone). Since fashion data constitutes a subset of these large corpora, these models possess inherent fashion retrieval capabilities. Therefore, we evaluate them directly on the U-FIRE benchmark without further fine-tuning.

    \item \textbf{Fashion-Domain VLP Models (Fashion-VLP):} This group of models builds upon fine-tuning vision–language models with fashion-specific data, including Fame-ViL~\cite{famevil}, FashionSAP~\cite{fashionsap}, and DA-Fashion~\cite{dafashion}. As their training data already cover the fashion domain, these models are evaluated directly on the U-FIRE benchmark without additional fine-tuning.   
    
    \item \textbf{Fine-tuned Models:} We further include two baseline models that are fine-tuned specifically on the U-FIRE training set, covering two mainstream retrieval paradigms: (1) UniIR$^*$~\cite{uniir}, which adopts a CLIP-L/14 backbone with score-level fusion and is optimized via full-parameter fine-tuning; and (2) the Qwen3-VL model (4B version), adapted using Low-Rank Adaptation (LoRA) with standard contrastive learning and learnable retrieval tokens. 
 
\end{itemize}

The comparison results are presented in Table \ref{tab:main_results}, from which we draw the following observations.
1) FashionLens consistently outperforms the baselines across most datasets, demonstrating its effectiveness for versatile fashion image retrieval. The performance gap between our method and GME is smallest on the FashionIQ datasets, likely because these datasets overlap with GME’s training data. It is noteworthy that GME employs a 7B backbone, whereas FashionLens uses only a 4B backbone, highlighting the efficiency of our approach. Meanwhile, although the Fashion-VLP model DA-Fashion underperforms on most other tasks, it achieves the best results on the FashionGen dataset. This is likely because FashionGen is included in DA-Fashion's training data, and Text$\to$Image retrieval is a dominant task in its optimization.
2) The fine-tuned UniIR$^*$ achieves substantial gains compared to the initial UniIR, which underscores the necessity of constructing a comprehensive fashion retrieval benchmark like U-FIRE.
3) Overall, MLLM-based methods achieve better performance than VLP-based counterparts, confirming that the MLLM architecture is inherently better suited for processing complex multimodal inputs and interpreting diverse retrieval instructions compared to VLP frameworks.
4) FashionLens significantly outperforms all baselines on both unseen tasks, demonstrating strong generalization and broad applicability in the fashion domain. 

\subsection{Ablation Study}
To verify the influence of each component in our model, we compare FashionLens with the following derivations:
\begin{itemize}[leftmargin=\parindent, labelsep=0.5em]
    \item \textbf{\#1. Base (Qwen3-VL-4B)}: This is the baseline model, which fine-tunes Qwen3-VL-4B via LoRA with learnable special tokens and contrastive learning. It uses the naive query representation $\mathbf{q}_0$ and random sampling.
    \item \textbf{\#2. Base+GGAS}: To investigate the impact of gradient-guided sampling, we replace the random sampling strategy in the base model with our GGAS strategy.
    \item \textbf{\#3. Proposal\_Only}: To demonstrate the necessity of performing Slerp to derive the final query representation $\mathbf{q}$,  we directly utilize the intention-oriented adaptation proposal $\mathbf{q}_p$ as $\mathbf{q}$.
    \item \textbf{\#4. Linear\_Interp}: To justify the effectiveness of Slerp, we replace it with standard linear interpolation in Eqn. (\ref{slerp}).
    \item \textbf{\#5. w/o\_Size\_Refine}: To investigate the importance of size-aware probability refinement, we set $\gamma=0$ in Eqn. (\ref{gamma}).
    \item \textbf{\#6. Shared\_Params.}: To assess the necessity of conducting instance-level proposal generation, we remove the MLP and directly treat the projection matrices $\mathbf{A}$, $\mathbf{B}$ and the interpolation coefficient $\lambda$ as learnable parameters shared across all samples. 
\end{itemize}

\begin{table}[t!]
    \centering
    \caption{Ablation study. We report the \textbf{Average Performance} (R@1, R@5, R@10, mR) averaged over all training tasks.}
    % \vspace{-1em}
    \label{tab:ablation_study}
    \setlength{\tabcolsep}{4pt} 
    \resizebox{\columnwidth}{!}{%
    \begin{tabular}{c|c|c|cccc} 
    \thickhline
    \multicolumn{3}{c|}{\textbf{Method}} & \multicolumn{4}{c}{\textbf{Average Performance}} \\ 
    \hline
    \textbf{\#} & \textbf{Query Rep.} & \textbf{Sampling Stra.} & \textbf{R@1} & \textbf{R@5} & \textbf{R@10} & \textbf{mR} \\
    \hline
    1 & $\mathbf{q}_0$ & Random & $22.62$ & $45.00$ & $53.61$ & $40.41$ \\
    2 & $\mathbf{q}_0$ & GGAS & $30.76$ & $54.09$ & $62.08$ & $48.98$ \\
    3 & $\mathbf{q}_{p}$ & GGAS & $31.08$ & $53.98$ & $61.90$ & $48.99$ \\
    4 & $\mathbf{q}$ (Linear) & GGAS & $31.00$ & $53.60$ & $61.27$ & $48.62$ \\
    5 & $\mathbf{q}$ (Slerp) & GGAS (w/o\_size) & $31.03$ & $54.31$ & $62.04$ & $49.12$ \\
    6 & $\mathbf{q}$ (shared params.) & GGAS & $31.41$ & $54.91$ & $62.36$ & $49.56$ \\
    \hline
    Ours & $\mathbf{q}$ (Slerp) & GGAS & \textbf{$32.17$} & \textbf{$55.21$} & \textbf{$62.65$} & \textbf{$50.01$} \\
    \thickhline
    \end{tabular}%
    }
\end{table}

\begin{figure}[tp]
	\includegraphics[width=\linewidth]{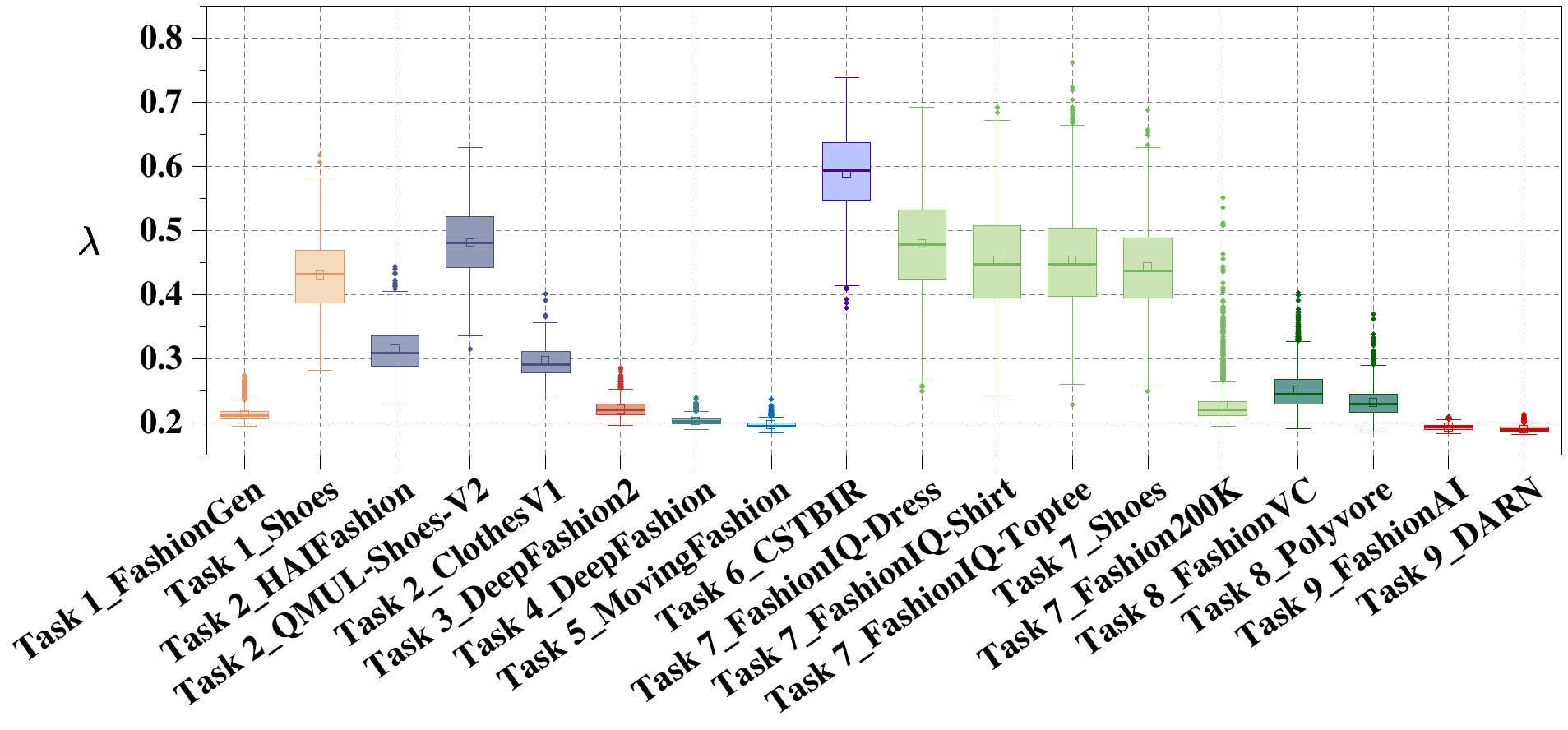}
    \vspace{-1.1em}
	\caption{Distribution of the learnable interpolation weight $\lambda$ across different test sets.}
	\label{fig:lambda}
    \vspace{-1em}
\end{figure}

Table~\ref{tab:ablation_study} shows the average performance of different variant methods over all training tasks.
From Table~\ref{tab:ablation_study}, we gain the following observations. 
1) The most substantial performance improvement stems from our GGAS strategy  ($\#2$). 
This confirms that the inherent data imbalance and varying task difficulties in versatile retrieval severely hinder optimization. GGAS effectively mitigates this by utilizing gradient norms to prioritize ``hard'' tasks. 
Directly using the adaptation proposal $\mathbf{q}_p$ ($\#3$) performs worse than our FashionLens, implying that dynamically retaining semantic information from the original query $\mathbf{q}_0$ is vital for achieving precise query representation adaptation.
3) Calibration via Slerp outperforms standard linear interpolation ($\#4$), likely because Euclidean interpolation can distort feature magnitudes on the unit hypersphere, whereas Slerp enables controllable adaptation of representations through a pure rotation. 
4) The incorporation of size-aware probability refinement ($\#5$) does help stabilize the multi-task optimization process and improve performance.
5) Instance level parameterization provides superior adaptivity compared to shared parameters ($\#6$),  indicating that different query samples do require unique adaptation directions and magnitudes. 

\subsection{Visualization and Analysis}
To gain deeper insights into our model, we visualize the intermediate results of our two key components.

\textbf{On PGSQC}. To provide insight into how FashionLens adapts queries, we visualize the distribution of the interpolation coefficient $\lambda$, which controls the magnitude of query adaptation, across different test sets in Figure~\ref{fig:lambda}. We observe a negative correlation between training data scale and $\lambda$, even within the same task categories. For instance, in Text$\to$Image (Task 1), the large-scale FashionGen (49k samples) exhibits a low $\lambda$, whereas the smaller Shoes dataset (2.4k samples) yields a much higher $\lambda$. A similar trend is observed in Task 7 between Fashion200k (47k, low $\lambda$) and FashionIQ ($\sim$5k, high $\lambda$). This pattern is expected: when data is abundant, the corresponding query-target matching patterns are more likely to be captured by the initial query representation produced by the unified backbone ($\mathbf{q}_0$), as the task tends to dominate model training, requiring only minor calibration. In contrast, in data-scarce scenarios, task-specific patterns are less well represented by $\mathbf{q}_0$, necessitating a larger $\lambda$ for further adaptation of the query representation.

\begin{figure}[tp]
	\includegraphics[width=\linewidth]{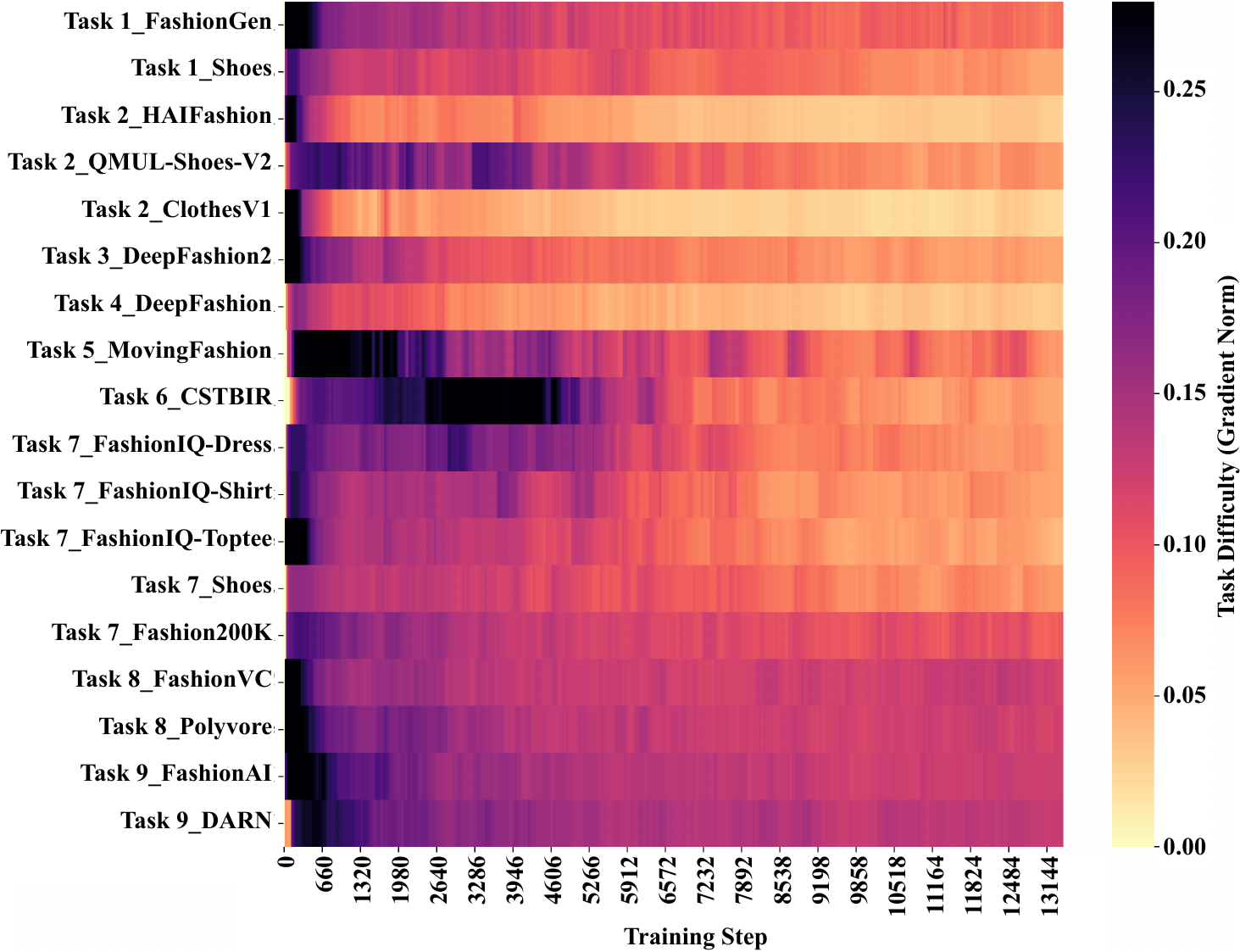}
    \vspace{-1.2em}
	\caption{Visualization of task difficulty measured by gradient norms during training.}
	\label{fig:difficulty}
    \vspace{-1.2em}
\end{figure}

\textbf{On GGAS}.  We further visualize the evolution of task difficulty, measured by gradient norms, for each dataset throughout the training process in Figure~\ref{fig:difficulty}. First, the heatmaps show a general downward trend in gradient norms across all datasets as training progresses, which is expected as the model gradually converges. Second, the rate of gradient decay differs across tasks, confirming that diverse fashion retrieval tasks have inherently different levels of optimization complexity. Third, the heatmaps reveal a clear stratification of task difficulty: tasks involving abstract reasoning or complex cross-modal interactions, such as compatibility matching (Task 8) and attribute-conditional matching (Task 9), consistently exhibit higher gradient norms compared to simpler tasks like In-Shop Retrieval (Task 4). This reflects the effectiveness of using gradient norms as a measure of task difficulty, as more challenging tasks naturally correspond to larger gradients.

%% file: 6_conclusion.tex
\section{Conclusion}
In this paper, we addressed the limitations of fragmented fashion retrieval by proposing FashionLens, a unified framework for versatile fashion image retrieval. We established U-FIRE, a comprehensive benchmark unifying $15$ datasets to facilitate standardized evaluation. To handle diverse retrieval intents, we introduced the Proposal-Guided Spherical Query Calibrator (PGSQC), which dynamically calibrates query representations on the hypersphere. Additionally, we designed Gradient-Guided Adaptive Sampling (GGAS) to resolve optimization imbalances by harmonizing task difficulty with dataset scale. Experimental results demonstrate that FashionLens not only outperforms existing methods on standard tasks but also generalizes to complex, unseen scenarios. We hope that this work will encourage the community to move beyond fragmented solutions and advance the development of versatile fashion image retrieval systems.

%% file: 0_main.bbl
\begin{thebibliography}{10}
\providecommand{\url}[1]{#1}
\csname url@samestyle\endcsname
\providecommand{\newblock}{\relax}
\providecommand{\bibinfo}[2]{#2}
\providecommand{\BIBentrySTDinterwordspacing}{\spaceskip=0pt\relax}
\providecommand{\BIBentryALTinterwordstretchfactor}{4}
\providecommand{\BIBentryALTinterwordspacing}{\spaceskip=\fontdimen2\font plus
\BIBentryALTinterwordstretchfactor\fontdimen3\font minus \fontdimen4\font\relax}
\providecommand{\BIBforeignlanguage}[2]{{%
\expandafter\ifx\csname l@#1\endcsname\relax
\typeout{** WARNING: IEEEtran.bst: No hyphenation pattern has been}%
\typeout{** loaded for the language `#1'. Using the pattern for}%
\typeout{** the default language instead.}%
\else
\language=\csname l@#1\endcsname
\fi
#2}}
\providecommand{\BIBdecl}{\relax}
\BIBdecl

\bibitem{SongFHYLN18}
X.~Song, F.~Feng, X.~Han, X.~Yang, W.~Liu, and L.~Nie, ``Neural compatibility modeling with attentive knowledge distillation,'' in \emph{Proceedings of the International {ACM} {SIGIR} Conference on Research and Development in Information Retrieval}.\hskip 1em plus 0.5em minus 0.4em\relax {ACM}, 2018, pp. 5--14.

\bibitem{DBLP:journals/tmm/JingCGNS23}
P.~Jing, K.~Cui, W.~Guan, L.~Nie, and Y.~Su, ``Category-aware multimodal attention network for fashion compatibility modeling,'' \emph{IEEE Transactions on Multimedia}, vol.~25, pp. 9120--9131, 2023.

\bibitem{DBLP:journals/tmm/GuWSPCK19}
X.~Gu, Y.~Wong, L.~Shou, P.~Peng, G.~Chen, and M.~S. Kankanhalli, ``Multi-modal and multi-domain embedding learning for fashion retrieval and analysis,'' \emph{IEEE Transactions on Multimedia}, vol.~21, no.~6, pp. 1524--1537, 2019.

\bibitem{11488029}
L.~Xiao and T.~Yamasaki, ``A multihead continual learning framework for fine-grained fashion image retrieval with contrastive learning and exponential moving average distillation,'' \emph{IEEE Transactions on Multimedia}, pp. 1--10, 2026.

\bibitem{QMUL_Shoe_V2}
Q.~Yu, F.~Liu, Y.~Song, T.~Xiang, T.~M. Hospedales, and C.~C. Loy, ``Sketch me that shoe,'' in \emph{{IEEE} Conference on Computer Vision and Pattern Recognition}.\hskip 1em plus 0.5em minus 0.4em\relax {IEEE} Computer Society, 2016, pp. 799--807.

\bibitem{Deepfashion}
Z.~Liu, P.~Luo, S.~Qiu, X.~Wang, and X.~Tang, ``Deepfashion: Powering robust clothes recognition and retrieval with rich annotations,'' in \emph{{IEEE} Conference on Computer Vision and Pattern Recognition}.\hskip 1em plus 0.5em minus 0.4em\relax {IEEE} Computer Society, 2016, pp. 1096--1104.

\bibitem{MovingFashion}
M.~Godi, C.~Joppi, G.~Skenderi, and M.~Cristani, ``Movingfashion: a benchmark for the video-to-shop challenge,'' in \emph{Proceedings of the {IEEE/CVF} Winter Conference on Applications of Computer Vision}.\hskip 1em plus 0.5em minus 0.4em\relax {IEEE}, 2022, pp. 517--525.

\bibitem{fashionbert}
D.~Gao, L.~Jin, B.~Chen, M.~Qiu, P.~Li, Y.~Wei, Y.~Hu, and H.~Wang, ``Fashionbert: Text and image matching with adaptive loss for cross-modal retrieval,'' in \emph{Proceedings of the International {ACM} {SIGIR} Conference on Research and Development in Information Retrieval}.\hskip 1em plus 0.5em minus 0.4em\relax {ACM}, 2020, pp. 2251--2260.

\bibitem{famevil}
X.~Han, X.~Zhu, L.~Yu, L.~Zhang, Y.~Song, and T.~Xiang, ``Fame-vil: Multi-tasking vision-language model for heterogeneous fashion tasks,'' in \emph{Proceedings of the {IEEE} Conference on Computer Vision and Pattern Recognition}.\hskip 1em plus 0.5em minus 0.4em\relax {IEEE}, 2023, pp. 2669--2680.

\bibitem{fashionsap}
Y.~Han, L.~Zhang, Q.~Chen, Z.~Chen, Z.~Li, J.~Yang, and Z.~Cao, ``Fashionsap: Symbols and attributes prompt for fine-grained fashion vision-language pre-training,'' in \emph{Proceedings of the {IEEE/CVF} Conference on Computer Vision and Pattern Recognition}.\hskip 1em plus 0.5em minus 0.4em\relax {IEEE}, 2023, pp. 15\,028--15\,038.

\bibitem{clip}
A.~Radford, J.~W. Kim, C.~Hallacy, A.~Ramesh, G.~Goh, S.~Agarwal, G.~Sastry, A.~Askell, P.~Mishkin, J.~Clark, G.~Krueger, and I.~Sutskever, ``Learning transferable visual models from natural language supervision,'' in \emph{Proceedings of the International Conference on Machine Learning}.\hskip 1em plus 0.5em minus 0.4em\relax {PMLR}, 2021, pp. 8748--8763.

\bibitem{blip}
J.~Li, D.~Li, C.~Xiong, and S.~C.~H. Hoi, ``{BLIP:} bootstrapping language-image pre-training for unified vision-language understanding and generation,'' in \emph{Proceedings of the International Conference on Machine Learning}.\hskip 1em plus 0.5em minus 0.4em\relax {PMLR}, 2022, pp. 12\,888--12\,900.

\bibitem{uniir}
C.~Wei, Y.~Chen, H.~Chen, H.~Hu, G.~Zhang, J.~Fu, A.~Ritter, and W.~Chen, ``Uniir: Training and benchmarking universal multimodal information retrievers,'' in \emph{Proceedings of the European Conference on Computer Vision}.\hskip 1em plus 0.5em minus 0.4em\relax Springer, 2024, pp. 387--404.

\bibitem{vlm2vec}
Z.~Jiang, R.~Meng, X.~Yang, S.~Yavuz, Y.~Zhou, and W.~Chen, ``Vlm2vec: Training vision-language models for massive multimodal embedding tasks,'' in \emph{Proceedings of the International Conference on Learning Representations}.\hskip 1em plus 0.5em minus 0.4em\relax OpenReview.net, 2025.

\bibitem{VLM2VECV2}
R.~Meng, Z.~Jiang, Y.~Liu, M.~Su, X.~Yang, Y.~Fu, C.~Qin, Z.~Chen, R.~Xu, C.~Xiong, Y.~Zhou, W.~Chen, and S.~Yavuz, ``Vlm2vec-v2: Advancing multimodal embedding for videos, images, and visual documents,'' \emph{CoRR}, vol. abs/2507.04590, 2025.

\bibitem{negative1}
S.~Liu, Y.~Liang, and A.~Gitter, ``Loss-balanced task weighting to reduce negative transfer in multi-task learning,'' in \emph{Proceedings of the {AAAI} Conference on Artificial Intelligence}.\hskip 1em plus 0.5em minus 0.4em\relax {AAAI} Press, 2019, pp. 9977--9978.

\bibitem{negative2}
J.~Zhou, Q.~Yu, C.~Luo, and J.~Zhang, ``Feature decomposition for reducing negative transfer: {A} novel multi-task learning method for recommender system (student abstract),'' in \emph{Proceedings of the {AAAI} Conference on Artificial Intelligence}.\hskip 1em plus 0.5em minus 0.4em\relax {AAAI} Press, 2023, pp. 16\,390--16\,391.

\bibitem{dafashion}
W.~Guan, K.~Wang, X.~Song, K.~Zhang, X.~Chang, and S.~Zhang, ``Dual alignment-enhanced fashion vision-language pre-training,'' \emph{ACM Transactions on Multimedia Computing, Communications and Applications}, just Accepted.

\bibitem{fashiongen}
N.~Rostamzadeh, S.~Hosseini, T.~Boquet, W.~Stokowiec, Y.~Zhang, C.~Jauvin, and C.~Pal, ``Fashion-gen: The generative fashion dataset and challenge,'' \emph{CoRR}, vol. abs/1806.08317, 2018.

\bibitem{FashionIQ}
H.~Wu, Y.~Gao, X.~Guo, Z.~Al{-}Halah, S.~Rennie, K.~Grauman, and R.~Feris, ``Fashion {IQ:} {A} new dataset towards retrieving images by natural language feedback,'' in \emph{{IEEE} Conference on Computer Vision and Pattern Recognition}.\hskip 1em plus 0.5em minus 0.4em\relax Computer Vision Foundation / {IEEE}, 2021, pp. 11\,307--11\,317.

\bibitem{gme}
X.~Zhang, Y.~Zhang, W.~Xie, M.~Li, Z.~Dai, D.~Long, P.~Xie, M.~Zhang, W.~Li, and M.~Zhang, ``Bridging modalities: Improving universal multimodal retrieval by multimodal large language models,'' in \emph{Proceedings of the {IEEE/CVF} Conference on Computer Vision and Pattern Recognition}.\hskip 1em plus 0.5em minus 0.4em\relax {IEEE}, 2025, pp. 9274--9285.

\bibitem{mmembed}
S.~Lin, C.~Lee, M.~Shoeybi, J.~Lin, B.~Catanzaro, and W.~Ping, ``Mm-embed: Universal multimodal retrieval with multimodal {LLMS},'' in \emph{Proceedings of the International Conference on Learning Representations}.\hskip 1em plus 0.5em minus 0.4em\relax OpenReview.net, 2025.

\bibitem{dqucir}
H.~Wen, X.~Song, X.~Chen, Y.~Wei, L.~Nie, and T.~Chua, ``Simple but effective raw-data level multimodal fusion for composed image retrieval,'' in \emph{Proceedings of the International {ACM} {SIGIR} Conference on Research and Development in Information Retrieval}.\hskip 1em plus 0.5em minus 0.4em\relax {ACM}, 2024, pp. 229--239.

\bibitem{DBLP:journals/tmm/DingMWC22}
Y.~Ding, Y.~Ma, W.~K. Wong, and T.~Chua, ``Modeling instant user intent and content-level transition for sequential fashion recommendation,'' \emph{IEEE Transactions on Multimedia}, vol.~24, pp. 2687--2700, 2022.

\bibitem{DBLP:journals/tmm/LuHYCZ23}
Z.~Lu, Y.~Hu, C.~Yu, Y.~Chen, and B.~Zeng, ``Learning fashion compatibility with context conditioning embedding,'' \emph{IEEE Transactions on Multimedia}, vol.~25, pp. 5516--5526, 2023.

\bibitem{Shoes}
X.~Guo, H.~Wu, Y.~Cheng, S.~Rennie, G.~Tesauro, and R.~S. Feris, ``Dialog-based interactive image retrieval,'' in \emph{Advances in Neural Information Processing Systems}, 2018, pp. 676--686.

\bibitem{HAIFashion}
J.~Jiang, X.~Li, W.~Yu, and D.~Wu, ``{HAIFIT:} human-centered {AI} for fashion image translation,'' \emph{CoRR}, vol. abs/2403.08651, 2024.

\bibitem{Clothes_V1}
J.~Jiang, D.~Wu, Z.~Jiang, and W.~Yu, ``Simple yet efficient: Towards self-supervised {FG-SBIR} with unified sample feature alignment,'' \emph{CoRR}, vol. abs/2406.11551, 2024.

\bibitem{Deepfashion2}
Y.~Ge, R.~Zhang, X.~Wang, X.~Tang, and P.~Luo, ``Deepfashion2: {A} versatile benchmark for detection, pose estimation, segmentation and re-identification of clothing images,'' in \emph{Proceedings of the {IEEE} Conference on Computer Vision and Pattern Recognition}.\hskip 1em plus 0.5em minus 0.4em\relax Computer Vision Foundation / {IEEE}, 2019, pp. 5337--5345.

\bibitem{cstbir}
P.~Gatti, K.~Parikh, D.~P. Paul, M.~Gupta, and A.~Mishra, ``Composite sketch+text queries for retrieving objects with elusive names and complex interactions,'' in \emph{{AAAI} Conference on Artificial Intelligence}.\hskip 1em plus 0.5em minus 0.4em\relax {AAAI} Press, 2024, pp. 1869--1877.

\bibitem{Fashion200k}
X.~Han, Z.~Wu, P.~X. Huang, X.~Zhang, M.~Zhu, Y.~Li, Y.~Zhao, and L.~S. Davis, ``Automatic spatially-aware fashion concept discovery,'' in \emph{{IEEE} International Conference on Computer Vision}.\hskip 1em plus 0.5em minus 0.4em\relax {IEEE} Computer Society, 2017, pp. 1472--1480.

\bibitem{fashionvc}
X.~Song, F.~Feng, J.~Liu, Z.~Li, L.~Nie, and J.~Ma, ``Neurostylist: Neural compatibility modeling for clothing matching,'' in \emph{Proceedings of the {ACM} on Multimedia Conference}.\hskip 1em plus 0.5em minus 0.4em\relax {ACM}, 2017, pp. 753--761.

\bibitem{polyvore}
X.~Han, Z.~Wu, Y.~Jiang, and L.~S. Davis, ``Learning fashion compatibility with bidirectional lstms,'' in \emph{Proceedings of the {ACM} on Multimedia Conference}.\hskip 1em plus 0.5em minus 0.4em\relax {ACM}, 2017, pp. 1078--1086.

\bibitem{FashionAI}
X.~Zou, X.~Kong, W.~Wong, C.~Wang, Y.~Liu, and Y.~Cao, ``Fashionai: {A} hierarchical dataset for fashion understanding,'' in \emph{Proceedings of the {IEEE} Conference on Computer Vision and Pattern Recognition Workshops}.\hskip 1em plus 0.5em minus 0.4em\relax Computer Vision Foundation / {IEEE}, 2019, pp. 296--304.

\bibitem{DARN}
J.~Huang, R.~S. Feris, Q.~Chen, and S.~Yan, ``Cross-domain image retrieval with a dual attribute-aware ranking network,'' in \emph{Proceedings of the {IEEE} International Conference on Computer Vision}.\hskip 1em plus 0.5em minus 0.4em\relax {IEEE} Computer Society, 2015, pp. 1062--1070.

\bibitem{qwen3vl}
Q.~Team, ``Qwen3-vl technical report,'' \emph{CoRR}, vol. abs/2511.21631, 2025.

\bibitem{lora}
E.~J. Hu, Y.~Shen, P.~Wallis, Z.~Allen{-}Zhu, Y.~Li, S.~Wang, L.~Wang, and W.~Chen, ``Lora: Low-rank adaptation of large language models,'' in \emph{Proceedings of the International Conference on Learning Representations}.\hskip 1em plus 0.5em minus 0.4em\relax OpenReview.net, 2022.

\bibitem{adamw}
I.~Loshchilov and F.~Hutter, ``Decoupled weight decay regularization,'' in \emph{Proceedings of the International Conference on Learning Representations}.\hskip 1em plus 0.5em minus 0.4em\relax OpenReview.net, 2019.

\bibitem{gradcache}
L.~Gao, Y.~Zhang, J.~Han, and J.~Callan, ``Scaling deep contrastive learning batch size under memory limited setup,'' in \emph{Proceedings of the Workshop on Representation Learning for NLP}.\hskip 1em plus 0.5em minus 0.4em\relax Association for Computational Linguistics, 2021, pp. 316--321.

\end{thebibliography}
